\documentclass[11pt,a4paper]{article}

\usepackage[top=1in,bottom=1in,left=1.25in,right=1.25in]{geometry}

\usepackage[utf8]{inputenc}
\usepackage[T1]{fontenc}
\usepackage{times}
\usepackage{microtype}
\usepackage{setspace}
\usepackage{amsmath,amsfonts,amssymb}
\usepackage{nicefrac}

\usepackage{booktabs}
\usepackage{array}
\usepackage{multirow}
\usepackage{tabularx}
\usepackage{colortbl}
\usepackage{xcolor}
\definecolor{headerblue}{RGB}{30,90,160}
\definecolor{rowgray}{RGB}{245,245,248}
\definecolor{darkblue}{RGB}{20,60,120}

\usepackage{graphicx}
\usepackage{float}
\usepackage{caption}
\usepackage{subcaption}
\usepackage{tikz}
\usepackage{pgfplots}
\pgfplotsset{compat=1.18}

\usepackage{enumitem}

\usepackage{mdframed}
\mdfdefinestyle{abstractstyle}{
  linecolor=darkblue,
  linewidth=0.8pt,
  topline=true,
  bottomline=true,
  leftline=false,
  rightline=false,
  innertopmargin=8pt,
  innerbottommargin=8pt,
  innerleftmargin=10pt,
  innerrightmargin=10pt,
  backgroundcolor=white
}
\mdfdefinestyle{quotestyle}{
  linecolor=darkblue!40,
  linewidth=0.5pt,
  topline=false,
  bottomline=false,
  leftline=true,
  rightline=false,
  innertopmargin=6pt,
  innerbottommargin=6pt,
  innerleftmargin=12pt,
  innerrightmargin=6pt,
  backgroundcolor=rowgray
}

\usepackage{pifont}
\newcommand{\cmark}{\ding{51}}
\newcommand{\xmark}{\ding{55}}

\usepackage{titlesec}
\titleformat{\section}{\large\bfseries\color{darkblue}}{\thesection}{1em}{}[\titlerule]
\titleformat{\subsection}{\normalsize\bfseries\color{darkblue}}{\thesubsection}{1em}{}
\titleformat{\subsubsection}{\normalsize\bfseries}{\thesubsubsection}{1em}{}

\usepackage[numbers,sort&compress]{natbib}
\usepackage{hyperref}
\hypersetup{
    pdftitle={SenseAI: A Human-in-the-Loop Dataset for RLHF-Aligned Financial Sentiment Reasoning},
    pdfauthor={Berny Kabalisa},
    colorlinks=true,
    linkcolor=darkblue,
    citecolor=darkblue,
    urlcolor=darkblue
}

\usepackage{fancyhdr}
\title{%
  \LARGE\bfseries\color{darkblue}
  SenseAI: A Human-in-the-Loop Dataset for\\[4pt]
  RLHF-Aligned Financial Sentiment Reasoning
}

\author{%
  \textbf{Berny Kabalisa}\\[4pt]
  RizqSpark $\mid$ SenseAI\\[2pt]
  \href{mailto:bernykabalisa18@gmail.com}{\texttt{bernykabalisa18@gmail.com}}
}

\date{2026}

\begin{document}

\maketitle
\thispagestyle{empty}

\begin{mdframed}[style=abstractstyle]
\noindent\textbf{Abstract.}\quad
We introduce \textbf{SenseAI}, a proprietary, continuously collected corpus of
human-in-the-loop (HITL) validated financial sentiment data, enriched with
AI-generated reasoning chains, expert correction signals, and real-world market
outcome validation. The dataset is designed to address structural gaps in existing
financial NLP resources, specifically the absence of reasoning-aware annotations
and the lack of RLHF-compatible structure in widely used benchmarks such as
FinancialPhraseBank~\cite{malo2014good}.

Spanning \textbf{1,439 labelled data points} across \textbf{40 US-listed equities}
and \textbf{13 financial data categories}, SenseAI captures not only sentiment
classifications but the full decision-making context that produces them, including
AI reasoning chains, confidence scores, human correction signals, and real market
price outcomes. This multi-dimensional schema is architecturally aligned with
Reinforcement Learning from Human Feedback (RLHF) training
paradigms~\cite{christiano2017deep}, enabling direct integration into modern LLM
fine-tuning pipelines without additional preprocessing.

Analysis of the dataset yields \textbf{six novel empirical findings} regarding LLM
behaviour in financial sentiment reasoning, including the first documented
observation of \textit{latent reasoning drift} in financial NLP tasks, systematic
confidence over-hedging, forward projection, and a measurable \textit{Goldilocks
Zone} of correctable model error. These findings collectively validate the dataset's
core thesis: that general-purpose language models operate in a predictable,
correctable error regime in financial reasoning tasks, making high-quality HITL
correction data both necessary and sufficient for targeted improvement through
fine-tuning.

We discuss implications for enterprise financial AI agent deployment and invite
commercial collaboration. Researchers and practitioners interested in dataset
access, licensing, or co-authorship are encouraged to contact
\href{mailto:bernykabalisa18@gmail.com}{\texttt{bernykabalisa18@gmail.com}}.
\end{mdframed}

\vspace{6pt}

\newpage
\tableofcontents
\newpage

\section{Introduction}
\label{sec:intro}

Large language models (LLMs) have demonstrated remarkable linguistic capability
across a broad range of general-domain tasks, from open-ended question answering
to code generation and multi-step reasoning. Yet their application to high-stakes
financial environments remains critically and systematically limited in ways that are
not immediately obvious from general benchmark performance. Financial language is
structurally distinct from general text in three important respects: tone is frequently
understated and context-dependent, the same phrase can carry opposing implications
depending on macroeconomic conditions, and the consequences of misclassification
are not merely informational but directly monetary.

The most widely used financial sentiment benchmarks, most notably
FinancialPhraseBank~\cite{malo2014good} with its 4,840 labelled sentences, were
designed to answer a simpler question than the one the current generation of
LLM-powered financial applications demands. They capture what sentiment is, not how
an expert reasons toward that classification, not where AI reasoning diverges from
expert judgment, and not whether the sentiment is validated by subsequent market
behaviour. This structural limitation was acceptable when the primary use case was
binary sentiment classification with traditional NLP models. It is not acceptable when
the use case is training agentic financial AI systems that must interpret a continuous
stream of financial information with accuracy approaching that of a trained analyst.

This paper introduces \textbf{SenseAI}, a continuously collected dataset specifically
designed to address these limitations. SenseAI is built around a human-in-the-loop
data engine in which AI-generated sentiment interpretations of financial news
headlines are systematically reviewed and corrected by a human financial expert.
Each correction event constitutes a high-signal training data point: a precise,
timestamped record of where AI financial reasoning diverges from expert judgment, in
what direction, and by how much. The dataset further enriches each data point with
the full AI reasoning chain, a model confidence score, and real market price data
four hours after each classification, providing an objective external signal for
sentiment quality validation.

This paper makes the following contributions:

\begin{itemize}[leftmargin=1.8em, itemsep=4pt]
  \item We introduce the SenseAI dataset --- the first continuously collected,
        HITL-validated financial sentiment corpus enriched with AI reasoning chains,
        expert correction signals, confidence scores, and real-world market outcome
        validation.
  \item We provide a detailed description of the collection methodology, schema design,
        and quality control protocol, including a documented self-consistency testing
        process maintaining a 90\% annotation consistency target.
  \item We present an analysis of SenseAI's structural alignment with established RLHF
        training frameworks, demonstrating that the dataset satisfies all three structural
        requirements identified in the foundational RLHF literature: human preference
        signals, correction annotations, and reasoning context.
  \item We report six novel empirical findings from early dataset analysis, including
        the first documented observation of \textit{latent reasoning drift} in financial
        sentiment analysis tasks, which is only detectable through chain-of-thought data.
  \item We discuss commercial applications of the dataset in enterprise financial AI
        agent deployment and invite collaboration from researchers and industry
        practitioners.
\end{itemize}

The remainder of this paper is structured as follows. Section~\ref{sec:background}
reviews relevant background. Section~\ref{sec:dataset} describes the SenseAI
dataset. Section~\ref{sec:comparison} compares SenseAI with existing benchmarks.
Section~\ref{sec:findings} presents preliminary findings.
Section~\ref{sec:commercial} discusses commercial applications.
Section~\ref{sec:discussion} provides broader discussion.
Section~\ref{sec:limitations} addresses limitations. Section~\ref{sec:conclusion}
concludes.

\section{Background and Related Work}
\label{sec:background}

\subsection{Financial Sentiment Analysis}

Financial sentiment analysis has been an active research area for over a decade,
motivated by the well-established relationship between textual financial information
and asset price movements~\cite{tetlock2007giving}. Early work relied on lexicon-based
approaches using financial wordlists such as the Loughran-McDonald Sentiment Word
Lists~\cite{loughran2011liability}, which demonstrated that domain-specific vocabulary
significantly outperformed general-purpose sentiment lexicons on financial text.

The introduction of FinancialPhraseBank~\cite{malo2014good} represented a major
advance, establishing the first large-scale expert-annotated financial sentiment
benchmark, classifying 4,840 financial news sentences as positive, negative, or
neutral. Subsequent work expanded coverage: the FiQA dataset~\cite{maia2018www}
introduced aspect-level financial sentiment combined with question answering, and
the FLUE benchmark suite~\cite{shah2022flue} consolidated multiple financial NLP
tasks. BloombergGPT~\cite{wu2023bloomberggpt} demonstrated the substantial gains
achievable by training on a domain-specific financial corpus, validating the central
premise that financial domain-specificity is a meaningful performance driver.

Despite this progress, existing datasets share a critical structural limitation in the
context of modern LLM training: they capture what sentiment \textit{is}, not how an
expert \textit{reasons} toward that classification. None were designed with RLHF
alignment in mind. SenseAI addresses this gap directly.

\subsection{Reinforcement Learning from Human Feedback}

RLHF has emerged as the dominant methodology for aligning LLMs with human
preferences. Christiano et al.\ (2017)~\cite{christiano2017deep} established the theoretical
foundation, demonstrating that human preference signals over model outputs
meaningfully improve model behaviour beyond what raw text training achieves. InstructGPT~\cite{ouyang2022training}
operationalised RLHF at scale, showing that models trained on human feedback data
significantly outperform much larger models on instruction-following tasks.

The key insight is that RLHF data requires three structural properties: (1)~human
preference signals indicating which outputs are preferred, (2)~correction annotations
showing what the preferred output looks like, and (3)~reasoning context enabling the
model to learn the pattern behind corrections rather than just the corrections
themselves. SenseAI provides all three by design. Constitutional
AI~\cite{bai2022constitutional} has further reinforced that high-quality preference data
over smaller volumes outperforms large volumes of low-quality signals in
domain-specific fine-tuning.

\subsection{The Data Wall and Domain-Specific Training}

Recent work has identified a ``data wall'' phenomenon~\cite{villalobos2022will} --- the
approaching exhaustion of high-quality public text data for large-scale LLM
pretraining. As frontier labs have consumed the majority of high-quality internet text,
academic corpora, and digitised books, the marginal return on additional pretraining
data is diminishing. Sutskever,~I.\ (2024)~\cite{sutskever2024} and subsequent industry commentary have
suggested that proprietary, domain-specific, structurally rich datasets represent the
primary source of competitive differentiation in the next generation of model
development --- not merely because they enable task-specific improvements, but
because they are structurally irreproducible.

SenseAI is positioned at this intersection: domain-specific, continuously generated,
and structurally rich beyond what public financial corpora provide.

\section{The SenseAI Dataset}
\label{sec:dataset}

\subsection{Overview and Motivation}

SenseAI is a continuously growing corpus of HITL-validated financial news sentiment
data, collected since December~1, 2024. The dataset is motivated by a specific gap
in the financial NLP landscape: the absence of RLHF-ready training data that captures
not only what expert financial analysts conclude about financial news, but how they
reason toward that conclusion and where AI reasoning diverges from expert judgment.

The core design insight is that the value of a training signal for RLHF lies not in its
volume but in its precision. A dataset of 1,439 expert-validated correction events,
each enriched with the AI's full reasoning chain, is more valuable for fine-tuning a
financial sentiment model than 100,000 simple sentiment labels, because it teaches
the model the reasoning pattern that leads to correct judgment rather than just the
correct label.

\subsection{Data Structure and Schema}

Each data point in the SenseAI dataset captures 13 fields, collectively constituting
the full cognitive context of a financial sentiment classification event.

\begin{table}[H]
\centering
\caption{SenseAI Dataset Schema}
\label{tab:schema}
\renewcommand{\arraystretch}{1.35}
\begin{tabularx}{\textwidth}{>{\bfseries\color{darkblue}}p{3.2cm} X}
\toprule
\rowcolor{headerblue}
\textcolor{white}{\textbf{Field}} & \textcolor{white}{\textbf{Description}} \\
\midrule
Ticker & Equity symbol of the company referenced in the news item \\
\rowcolor{rowgray}
Timestamp & Precise date and time of the news event (Eastern Time) \\
AI News Headline & Financial news headline as processed by the language model \\
\rowcolor{rowgray}
AI Sentiment & Model-generated classification (Bullish / Slightly Bullish / Neutral / Slightly Bearish / Bearish) \\
AI Reasoning & Full chain-of-thought reasoning paragraph supporting the classification \\
\rowcolor{rowgray}
AI Confidence Score & Model self-assessed certainty expressed as a percentage \\
Price at Call & Stock price at the moment of sentiment classification \\
\rowcolor{rowgray}
Price 4h Later & Actual market price four hours after classification (outcome validation) \\
HITL Edited & Binary flag: whether the human expert corrected the AI classification \\
\rowcolor{rowgray}
Human Sentiment & Expert-corrected sentiment label (where HITL Edited = Yes) \\
Edit Type & Correction severity: Category 0--3 \\
\rowcolor{rowgray}
News Paragraph & Full source news text fed to the model \\
LLM Version & GPT model version used for generation \\
\bottomrule
\end{tabularx}
\end{table}

This multi-dimensional schema captures not only the sentiment label but the full
cognitive context of each classification decision --- a structural property that
distinguishes SenseAI from all existing financial sentiment datasets. Whereas
FinancialPhraseBank provides a sentence and a three-category label, SenseAI
provides ten additional dimensions of information per data point.

\subsection{Collection Methodology}

Financial news paragraphs are collected from financial data sources and processed
by a language model, which is prompted to produce four outputs simultaneously:
(1)~a summarised news headline, (2)~a sentiment classification from the five-category
schema, (3)~a full chain-of-thought reasoning paragraph, and (4)~a confidence score.

Each output is then reviewed by a human financial expert --- a postgraduate student
in Finance and Investments with domain expertise in financial news interpretation ---
through the HITL validation process. Where the AI classification diverges from expert
assessment, the reviewer provides a corrected classification and records the edit type:

\begin{table}[H]
\centering
\caption{Edit Type Taxonomy}
\label{tab:edittypes}
\renewcommand{\arraystretch}{1.3}
\begin{tabular}{>{\bfseries}c p{3.5cm} p{7.5cm}}
\toprule
\rowcolor{headerblue}
\textcolor{white}{\textbf{Category}} &
\textcolor{white}{\textbf{Label}} &
\textcolor{white}{\textbf{Description}} \\
\midrule
0 & No correction   & AI classification accepted as-is \\
\rowcolor{rowgray}
1 & Minor correction & Small sentiment adjustment (e.g., Slightly Bullish $\to$ Bullish) \\
2 & Moderate correction & Meaningful sentiment shift (e.g., Neutral $\to$ Slightly Bullish) \\
\rowcolor{rowgray}
3 & Complete reversal & Sentiment direction inverted (e.g., Bullish $\to$ Bearish) \\
\bottomrule
\end{tabular}
\end{table}

\subsubsection{Collection Density and Pre-Automation Variability}

The current dataset reflects variable daily collection density inherent to the
pre-automation phase of a HITL annotation pipeline. As a single expert annotator
operating across 40 equities and 13 data categories, daily coverage varies with news
flow volume and annotation throughput. This variability does not introduce systematic
bias --- collection occurs consistently across all covered equities and categories, with
density variation reflecting exogenous news volume and annotator capacity rather than
selective sampling. Planned automation of the collection pipeline will standardise
daily throughput and eliminate density variability in future dataset versions.

\subsection{Quality Control Protocol}

Quality control is maintained through a self-consistency testing protocol in which the
human reviewer periodically re-evaluates previously labelled data points without
reference to the original annotation. A target consistency rate of 90\% is maintained
and verified on a rolling basis. Instances where the re-evaluation diverges from the
original annotation are flagged for review, and the original annotation is updated
where the fresh assessment provides a more defensible classification. This protocol
is adapted from standard inter-annotator agreement methodology and applied to the
single-annotator context.

\subsection{RLHF Structural Alignment}

The SenseAI dataset is architecturally aligned with RLHF training requirements along
four dimensions established in the foundational
literature~\cite{christiano2017deep,ouyang2022training}:

\begin{description}[leftmargin=2em, style=nextline,
                    font=\bfseries\color{darkblue}]

  \item[Human preference signals.]
    The HITL correction flag provides explicit human preference data at the data point
    level, indicating which AI outputs were acceptable (Category~0) and which required
    expert correction (Categories~1--3). This binary preference signal is the core input
    to reward model training in standard RLHF pipelines.

  \item[Correction annotations.]
    Each corrected data point records both the original AI classification and the
    human-corrected classification, providing the preference pair that reward model
    training requires. The edit type taxonomy enriches this signal by indicating
    correction magnitude.

  \item[Reasoning context.]
    Each data point includes the AI's full chain-of-thought reasoning process, enabling
    models to learn not just correct output labels but the reasoning patterns that lead
    to correct classification. This component is the most absent from existing financial
    sentiment datasets and the most critical for RLHF effectiveness in high-stakes
    domains.

  \item[Outcome validation.]
    Real-world market price data four hours after each classification provides an
    objective external signal for sentiment quality validation, independent of human
    judgment --- a form of data quality verification unique to financial applications and
    absent from all prior financial sentiment datasets.

\end{description}

This structural alignment means SenseAI data can be used directly in RLHF training
pipelines --- specifically in supervised fine-tuning and reward model training stages ---
without additional preprocessing or reformatting.

\subsection{Dataset Statistics: Early Snapshot}

The following statistics are derived from the current SenseAI dataset comprising
1,439 labelled data points collected between December 2025 and March 2026 across
40 US-listed equities. These figures represent early-stage collection and will expand
substantially as automated infrastructure scales daily throughput.

\subsubsection{Scale and Coverage}

The dataset contains 1,439 labelled data points at the time of this snapshot,
exceeding FinancialPhraseBank (4,840~sentences) at the individual structured data
point level while providing substantially richer metadata per record. Coverage spans
40 US-listed equities across six sectors:

\begin{itemize}[leftmargin=1.8em, itemsep=2pt]
  \item \textbf{Mega-Cap Technology:} AAPL, MSFT, GOOGL, NVDA, AMZN, META
  \item \textbf{Financial Services:} JPM, GS, BAC, MS, V, MA, C, WFC
  \item \textbf{Healthcare \& Pharmaceuticals:} LLY, MRK, PFE, JNJ, ABBV, UNH
  \item \textbf{Energy:} XOM, CVX, BP, NEE, TTE
  \item \textbf{Consumer:} WMT, COST, MCD, KO, PEP, NKE
  \item \textbf{Industrials \& Defence:} BA, CAT, LMT, RTX
\end{itemize}

\subsubsection{AI Sentiment Distribution}

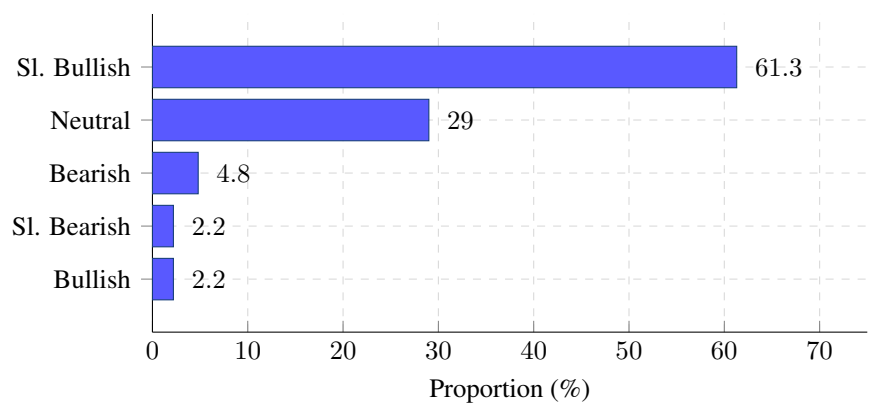
\begin{figure}[H]
\centering
\begin{tikzpicture}
\begin{axis}[
  xbar,
  width=0.75\textwidth,
  height=5.8cm,
  bar width=0.55cm,
  symbolic y coords={Bullish,Sl. Bearish,Bearish,Neutral,Sl. Bullish},
  ytick=data,
  xlabel={Proportion (\%)},
  xmin=0, xmax=75,
  xtick={0,10,20,30,40,50,60,70},
  nodes near coords,
  nodes near coords style={font=\small\bfseries},
  axis lines*=left,
  grid=major,
  grid style={dashed,gray!30},
  tick label style={font=\small},
  label style={font=\small},
  enlarge y limits=0.25,
  every node near coord/.append style={xshift=3pt}
]
\addplot[fill=blue!65, draw=darkblue] coordinates {
  (2.2,Bullish)
  (2.2,Sl. Bearish)
  (4.8,Bearish)
  (29.0,Neutral)
  (61.3,Sl. Bullish)
};
\end{axis}
\end{tikzpicture}
\caption{AI Sentiment Distribution ($n = 1{,}439$). The dominance of \textit{Slightly Bullish}
(61.3\%) relative to unhedged \textit{Bullish} (2.2\%) directly validates Finding~1:
systematic sentiment hypersensitivity to linguistic qualifiers.}
\label{fig:sentiment_bar}
\end{figure}

Analysis of AI-generated sentiment classifications reveals a pronounced distributional
skew toward hedged intermediate categories (Table~\ref{tab:sentiment_dist} and
Figure~\ref{fig:sentiment_bar}).

\begin{table}[H]
\centering
\caption{AI Sentiment Distribution ($n = 1{,}439$)}
\label{tab:sentiment_dist}
\renewcommand{\arraystretch}{1.3}
\begin{tabular}{lrr}
\toprule
\rowcolor{headerblue}
\textcolor{white}{\textbf{Sentiment Category}} &
\textcolor{white}{\textbf{Count}} &
\textcolor{white}{\textbf{Proportion}} \\
\midrule
Slightly Bullish  & 882 & 61.3\% \\
\rowcolor{rowgray}
Neutral           & 417 & 29.0\% \\
Bearish           &  69 &  4.8\% \\
\rowcolor{rowgray}
Slightly Bearish  &  32 &  2.2\% \\
Bullish           &  32 &  2.2\% \\
\midrule
\rowcolor{headerblue}
\textcolor{white}{\textbf{Total}} &
\textcolor{white}{\textbf{1,439}} &
\textcolor{white}{\textbf{100.0\%}} \\
\bottomrule
\end{tabular}
\end{table}

\subsubsection{HITL Correction Rate and Edit Type Distribution}

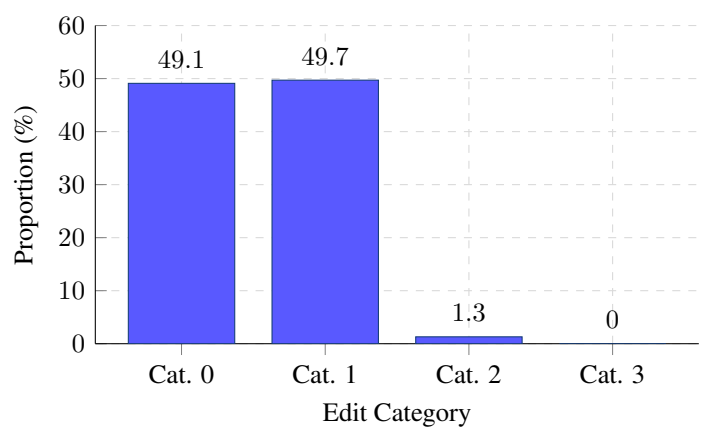
\begin{figure}[H]
\centering
\begin{tikzpicture}
\begin{axis}[
  ybar,
  width=0.65\textwidth,
  height=5.8cm,
  bar width=1.4cm,
  symbolic x coords={Cat. 0, Cat. 1, Cat. 2, Cat. 3},
  xtick=data,
  xlabel={Edit Category},
  ylabel={Proportion (\%)},
  ymin=0, ymax=60,
  ytick={0,10,20,30,40,50,60},
  nodes near coords,
  nodes near coords style={font=\small\bfseries},
  axis lines*=left,
  grid=major,
  grid style={dashed,gray!30},
  tick label style={font=\small},
  label style={font=\small},
  enlarge x limits=0.2,
  every node near coord/.append style={yshift=2pt}
]
\addplot[fill=blue!65, draw=darkblue] coordinates {
  (Cat. 0, 49.1)
  (Cat. 1, 49.7)
  (Cat. 2,  1.3)
  (Cat. 3,  0.0)
};
\end{axis}
\end{tikzpicture}
\caption{Edit type distribution. The near-symmetry between Category~0 (no edit,
49.1\%) and Category~1 (minor edit, 49.7\%), combined with the complete absence of
Category~3 reversals, defines the \textit{Goldilocks Zone} of correctable model error.}
\label{fig:edit_bar}
\end{figure}

Across the dataset, \textbf{51.4\%} of AI-generated sentiment classifications required
human expert correction, while 48.6\% were accepted without modification
(Table~\ref{tab:edit_types_dist} and Figure~\ref{fig:edit_bar}).

\begin{table}[H]
\centering
\caption{Edit Type Distribution ($n = 1{,}439$)}
\label{tab:edit_types_dist}
\renewcommand{\arraystretch}{1.3}
\begin{tabular}{clrr}
\toprule
\rowcolor{headerblue}
\textcolor{white}{\textbf{Category}} &
\textcolor{white}{\textbf{Description}} &
\textcolor{white}{\textbf{Count}} &
\textcolor{white}{\textbf{Proportion}} \\
\midrule
0 & No correction (accepted)  & 707 & 49.1\% \\
\rowcolor{rowgray}
1 & Minor correction          & 715 & 49.7\% \\
2 & Moderate correction       &  19 &  1.3\% \\
\rowcolor{rowgray}
3 & Complete reversal         &   0 &  0.0\% \\
\midrule
\rowcolor{headerblue}
  & \textcolor{white}{\textbf{Total}} &
    \textcolor{white}{\textbf{1,439}} &
    \textcolor{white}{\textbf{100.0\%}} \\
\bottomrule
\end{tabular}
\end{table}

\subsubsection{Confidence Score Distribution}

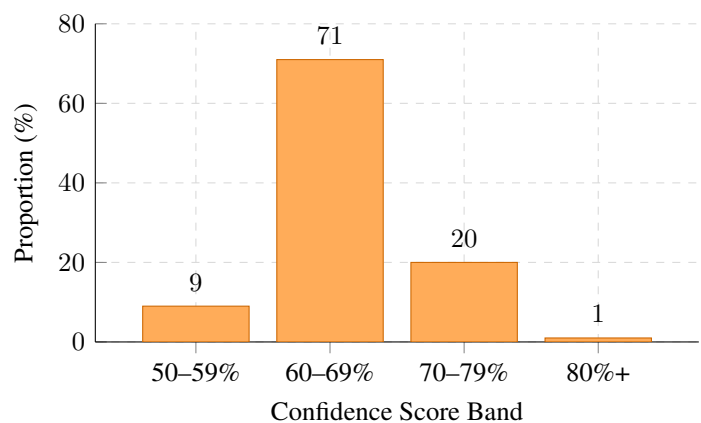
\begin{figure}[H]
\centering
\begin{tikzpicture}
\begin{axis}[
  ybar,
  width=0.65\textwidth,
  height=5.8cm,
  bar width=1.4cm,
  symbolic x coords={50--59\%, 60--69\%, 70--79\%, 80\%+},
  xtick=data,
  xlabel={Confidence Score Band},
  ylabel={Proportion (\%)},
  ymin=0, ymax=80,
  ytick={0,20,40,60,80},
  nodes near coords,
  nodes near coords style={font=\small\bfseries},
  axis lines*=left,
  grid=major,
  grid style={dashed,gray!30},
  tick label style={font=\small},
  label style={font=\small},
  enlarge x limits=0.25,
  every node near coord/.append style={yshift=2pt}
]
\addplot[fill=orange!65, draw=orange!80!black] coordinates {
  (50--59\%,  9.0)
  (60--69\%, 71.0)
  (70--79\%, 20.0)
  (80\%+,     1.0)
};
\end{axis}
\end{tikzpicture}
\caption{Confidence score distribution. The dominant concentration in the 60--69\%
band (71\% of all observations) directly validates Finding~2: systematic confidence
over-hedging regardless of signal clarity. No score exceeds 85\%.}
\label{fig:confidence_bar}
\end{figure}

AI confidence scores range from 55\% to 84\%, with a mean of \textbf{65.6\%}
($n = 1{,}439$). The distribution is heavily concentrated in the 60--69\% band
(Table~\ref{tab:confidence} and Figure~\ref{fig:confidence_bar}).

\begin{table}[H]
\centering
\caption{Confidence Score Distribution by Band}
\label{tab:confidence}
\renewcommand{\arraystretch}{1.3}
\begin{tabular}{lrr}
\toprule
\rowcolor{headerblue}
\textcolor{white}{\textbf{Confidence Band}} &
\textcolor{white}{\textbf{Count}} &
\textcolor{white}{\textbf{Proportion}} \\
\midrule
50--59\%  &  130 &  9.0\% \\
\rowcolor{rowgray}
60--69\%  & 1{,}022 & 71.0\% \\
70--79\%  &  288 & 20.0\% \\
\rowcolor{rowgray}
80\%+     &   14 &  1.0\% \\
\midrule
\rowcolor{headerblue}
\textcolor{white}{\textbf{Total}} &
\textcolor{white}{\textbf{1,439}} &
\textcolor{white}{\textbf{100.0\%}} \\
\midrule
\multicolumn{2}{l}{\textit{Mean}}  & \textit{65.6\%} \\
\multicolumn{2}{l}{\textit{Range}} & \textit{55--84\%} \\
\bottomrule
\end{tabular}
\end{table}

Critically, no statistically meaningful relationship is observed between confidence
score and HITL correction outcome. Data points with a 70\% confidence score are
corrected at approximately the same rate as data points with a 60\% confidence
score, confirming that the model's self-reported certainty is not a reliable predictor
of classification accuracy.

\subsubsection{Model Version Tracking}

SenseAI's timestamped collection methodology enables longitudinal tracking of model
version effects on output quality (Table~\ref{tab:model_versions}).

\begin{table}[H]
\centering
\caption{Model Version Distribution}
\label{tab:model_versions}
\renewcommand{\arraystretch}{1.3}
\begin{tabular}{lrr}
\toprule
\rowcolor{headerblue}
\textcolor{white}{\textbf{Model Version}} &
\textcolor{white}{\textbf{Approx.\ Data Points}} &
\textcolor{white}{\textbf{Proportion}} \\
\midrule
GPT-5.2    & 1{,}122 & 78\% \\
\rowcolor{rowgray}
GPT-5 mini &   216   & 15\% \\
GPT-5.1    &   101   &  7\% \\
\midrule
\rowcolor{headerblue}
\textcolor{white}{\textbf{Total}} &
\textcolor{white}{\textbf{1,439}} &
\textcolor{white}{\textbf{100\%}} \\
\bottomrule
\end{tabular}
\end{table}

The transition to GPT-5.2 produced a measurable reduction in Category~2 edit types,
which have declined to near-zero in recent collection cycles. This longitudinal tracking
capability --- absent from all static benchmark datasets --- enables SenseAI to serve as
a living instrument for measuring model improvement trajectories over time.

\section{Comparison with Existing Benchmarks}
\label{sec:comparison}

Table~\ref{tab:benchmark_comparison} presents a comparative analysis across nine
structural dimensions relevant to RLHF training.

\begin{table}[H]
\centering
\caption{Comparative Analysis: SenseAI vs.\ Existing Financial Sentiment Benchmarks}
\label{tab:benchmark_comparison}
\renewcommand{\arraystretch}{1.35}
\small
\begin{tabularx}{\textwidth}{>{\bfseries}p{3.6cm} X X X X}
\toprule
\rowcolor{headerblue}
\textcolor{white}{\textbf{Dimension}} &
\textcolor{white}{\textbf{SenseAI}} &
\textcolor{white}{\textbf{FinancialPhraseBank}} &
\textcolor{white}{\textbf{FiQA}} &
\textcolor{white}{\textbf{FLUE}} \\
\midrule
Dataset size          & 1,439+ (growing)   & 4,840 (static)    & 1,174 (static) & Multiple tasks \\
\rowcolor{rowgray}
Sentiment granularity & 5 categories       & 3 categories      & Continuous score & 3 categories \\
Reasoning chain       & \cmark\ Full CoT   & \xmark            & \xmark         & \xmark \\
\rowcolor{rowgray}
RLHF alignment        & \cmark\ Designed for & \xmark          & \xmark         & \xmark \\
Continuous collection & \cmark\ Daily      & \xmark            & \xmark         & \xmark \\
\rowcolor{rowgray}
Market outcome validation & \cmark\ 4h price & \xmark          & \xmark         & \xmark \\
Human correction signal & \cmark\ Expert HITL & \xmark         & \xmark         & \xmark \\
\rowcolor{rowgray}
Confidence score      & \cmark\ Per point  & \xmark            & \xmark         & \xmark \\
Model version tracking & \cmark\ Timestamped & \xmark          & \xmark         & \xmark \\
\rowcolor{rowgray}
Year created          & 2025 (ongoing)     & 2014              & 2018           & 2022 \\
\bottomrule
\end{tabularx}
\end{table}

The critical differentiator is not scale but structural depth. FinancialPhraseBank
provides three sentiment categories with no reasoning context and was designed in
2014 for a pre-RLHF era of NLP research. FiQA extends to continuous sentiment
scoring and aspect-level analysis but remains static and lacks correction signals. FLUE
consolidates multiple tasks but inherits the structural limitations of its component
datasets.

SenseAI provides five sentiment gradations, full reasoning chains, expert correction
annotations, confidence scoring, real-world outcome validation, and continuous
collection --- all in a single dataset designed from the ground up for RLHF. The
architectural gap reflects a fundamental shift in what the field requires: training data
for LLMs deployed as enterprise agents must enable models to learn reasoning
patterns, not just output labels. SenseAI is the first financial sentiment dataset
designed for this requirement.

\section{Preliminary Findings}
\label{sec:findings}

\subsection{Overview}

Analysis of the initial SenseAI collection period reveals six distinct and empirically
grounded findings regarding LLM behaviour in financial sentiment reasoning. These
findings collectively validate the core thesis of this paper: that general-purpose
language models operate in a ``Goldilocks Zone'' of financial reasoning --- sufficiently
accurate to be useful, yet sufficiently imperfect that human-in-the-loop supervision
and domain-specific fine-tuning provide measurable and commercially significant value.

Several of these findings are only detectable through the chain-of-thought reasoning
data that SenseAI captures. They would be invisible to any dataset recording only
sentiment labels --- which is itself a demonstration of the dataset's methodological
value.

\subsection{Finding 1: Sentiment Hypersensitivity to Linguistic Qualifiers}

A primary finding is that the model does not produce binary bullish or bearish
classifications. It consistently gravitates toward hedged intermediate classifications
even in the presence of strong directional signals. This behaviour is driven by
hypersensitivity to linguistic qualifiers within the news text.

For example, phrases such as ``strong revenues despite market conditions'' produce a
\textit{Slightly Bullish} rather than \textit{Bullish} output, as the model weights the
qualifier ``despite market conditions'' against the positive revenue signal. This pattern
is consistent across equities, sectors, and news types --- suggesting a structural
property of the model's alignment training rather than a case-specific response.

A model that systematically softens strong directional signals will underperform in
momentum-sensitive trading and signal-generation applications where the distinction
between Bullish and Slightly Bullish has direct position-sizing implications. The
predominance of intermediate classifications (61.3\% Slightly Bullish) combined with
the complete absence of Category~3 reversals confirms that this is a systematic
calibration bias, not random error.

\subsection{Finding 2: Systematic Confidence Over-Hedging}

Confidence scores cluster consistently in the 60--80\% range, with 71\% of all scores
falling in the 60--69\% band regardless of the directional clarity of the underlying
news item. We term this \textit{confidence over-hedging} --- a systematic understatement
of certainty that appears to be an artefact of general-purpose alignment training
rather than genuine uncertainty about the financial signal.

Critically, confidence scores are not reliably calibrated to accuracy. Data points with
a 70\% confidence score are corrected at approximately the same rate as data points
with a 60\% confidence score, confirming that the model's self-reported certainty does
not predict classification quality. Enterprise deployment pipelines that route
high-confidence outputs for automatic processing and low-confidence outputs for
human review would experience no meaningful quality benefit from this
threshold-based approach using current model confidence scores.

\subsection{Finding 3: Latent Reasoning Drift and Model Contamination}

A novel finding of significant theoretical and practical importance is that when
prompted to generate sentiment based solely on a given news item, the model
produces outputs influenced by factors not present in the provided text. Analysis of
reasoning chain data reveals that the model implicitly incorporates global context ---
learned associations between company names, historical performance patterns, and
broader market conditions --- even when explicitly absent from the input.

We term this phenomenon \textbf{\textit{latent reasoning drift}} --- the contamination of
single-document sentiment analysis by the model's internally learned multi-factor
reasoning patterns. This is an emergent property of large-scale pretraining on
globally correlated financial text. The model cannot fully isolate its sentiment
assessment to the specific information in the provided news item.

\begin{mdframed}[style=quotestyle]
\textit{This phenomenon has not, to our knowledge, been previously documented
empirically in financial sentiment analysis tasks. It is only detectable through
chain-of-thought reasoning data. Standard sentiment datasets that record only
classification labels cannot detect latent reasoning drift, because the drift occurs
within the reasoning process, not in the output label itself.}
\end{mdframed}

The SenseAI dataset therefore enables a novel line of research into the boundary
between grounded analysis and model contamination in financial NLP.

\subsection{Finding 4: The Goldilocks Zone --- Quantifying the Human Correction Rate}

Across the dataset, 51.4\% of classifications required expert correction, distributed
such that 49.7\% of all observations are Category~1 light corrections and 0.0\% are
Category~3 complete reversals. This confirms the model operates in the
\textbf{\textit{Goldilocks Zone}} of financial reasoning: accurate enough to provide
meaningful signal, imperfect enough that systematic human oversight provides
consistent corrective value.

\begin{mdframed}[style=quotestyle]
\textit{``We observe that approximately 50\% of AI-predicted sentiments require minor
human correction (Category~1), indicating that the model is largely accurate in
direction but consistently requires calibration in magnitude. Category~3 errors
(complete reversals) are entirely absent from the dataset, highlighting model
stability. This distribution characterises a Goldilocks Zone of correctable model
imperfection --- precisely the regime in which RLHF-based fine-tuning provides
maximum training signal relative to annotation cost.''}
\end{mdframed}

This characterisation has direct commercial significance: a fine-tuned model trained
on SenseAI correction data does not need to overcome catastrophic reasoning
failures --- it needs to close the gap on the consistent, predictable Category~1 hedging
errors that the current model produces at scale.

\subsection{Finding 5: Forward Projection and Grounded Reasoning Drift}

Reasoning chain analysis reveals a consistent pattern of temporal overreach: the
model injects assumptions about future events, outcomes, and market conditions not
present in the source text. We term this \textbf{\textit{forward projection}} --- a form of
reasoning that extends beyond the grounded analytical task into speculative territory.

Forward projection is only measurable through SenseAI's reasoning chain data. A
classification of ``Slightly Bullish'' generated through grounded analysis and one
generated through forward projection produce identical output labels, but they have
fundamentally different epistemological bases and different reliability profiles.
SenseAI therefore enables a novel line of research into the boundary between
grounded analysis and model hallucination in financial NLP.

\subsection{Finding 6: Model Version Effects on Error Rate}

Since the introduction of GPT-5.2, Category~2 errors (moderate corrections requiring
meaningful sentiment adjustment) have declined to near-zero in the dataset. This
improvement in base model quality does not eliminate the value of HITL supervision;
rather, it shifts the correction distribution further toward Category~1 light errors,
confirming that the Goldilocks Zone persists across model generations.

SenseAI is therefore not a dataset that becomes obsolete as models improve. Instead,
it becomes a longitudinal instrument for tracking and correcting the evolving error
profile of deployed financial AI systems --- a capability unique to continuously
collected datasets.

\subsection{Summary of Findings}

The six findings collectively establish three conclusions:

\begin{enumerate}[leftmargin=2em, itemsep=4pt]
  \item General-purpose LLMs exhibit consistent, predictable failure modes in financial
        sentiment reasoning that are correctable through targeted fine-tuning on
        domain-specific RLHF data.
  \item These failure modes --- particularly latent reasoning drift and forward projection
        --- are only fully detectable through reasoning chain data and are invisible to
        datasets recording sentiment labels alone, directly validating SenseAI's
        architectural approach.
  \item The Goldilocks Zone characterisation holds across model versions, ensuring the
        long-term relevance of the SenseAI dataset and methodology.
\end{enumerate}

\section{Commercial Applications}
\label{sec:commercial}

\subsection{The Enterprise Agent Opportunity}

The commercial significance of SenseAI is grounded in a fundamental shift in how AI
is being deployed in financial services. Agentic AI systems --- autonomous models
that continuously process information and generate actionable outputs without human
intervention --- are increasingly deployed at the enterprise level across financial
institutions, asset managers, hedge funds, and trading operations.

Unlike consumer AI applications where occasional inaccuracies are tolerable,
enterprise financial agents operate in high-stakes environments where sentiment
misclassification carries direct monetary consequences. A financial agent that
misreads the tone of an earnings call, misattributes sector-level bearish sentiment to
a specific equity, or fails to detect hedged language in a regulatory filing creates
measurable financial and reputational risk. The six findings documented in
Section~\ref{sec:findings} demonstrate that current general-purpose LLMs exhibit
exactly these failure modes systematically.

\medskip
\noindent\fbox{\parbox{0.965\textwidth}{%
\small\centering
Financial institutions, AI laboratories, and technology vendors interested in early
access to the SenseAI dataset for enterprise financial AI development are encouraged
to contact
\href{mailto:bernykabalisa18@gmail.com}{\texttt{bernykabalisa18@gmail.com}}.
}}

\subsection{Fine-Tuning Financial Models with SenseAI}

The SenseAI dataset is designed as the training foundation for fine-tuning language
models for financial sentiment reasoning tasks, improving performance across three
dimensions:

\begin{description}[leftmargin=2em, style=nextline,
                    font=\bfseries\color{darkblue}]

  \item[Sentiment accuracy.]
    Reducing misclassification rates benchmarked against FinancialPhraseBank and
    FiQA evaluation sets. The Goldilocks Zone finding suggests that even modest
    fine-tuning on SenseAI correction data will produce measurable improvement, as
    the model's error distribution is concentrated in a correctable, learnable regime.

  \item[Reasoning quality.]
    Improving the coherence and accuracy of model-generated financial reasoning
    chains, assessed through expert evaluation. Fine-tuning on SenseAI's reasoning
    chain data enables the model to learn not just correct output labels but the
    reasoning patterns that expert analysts employ.

  \item[Confidence calibration.]
    Improving the alignment between model confidence scores and actual classification
    accuracy. Finding~2 documents that current confidence scores have essentially no
    predictive relationship with classification quality. Fine-tuning on SenseAI data
    provides the training signal needed to produce calibrated confidence outputs
    actually useful for enterprise deployment thresholding.

\end{description}

A fine-tuned model trained on SenseAI data can be deployed directly within
enterprise financial AI infrastructure as the reasoning layer of an agentic system.
Researchers and AI laboratories interested in collaborative fine-tuning research are
invited to reach out at
\href{mailto:bernykabalisa18@gmail.com}{\texttt{bernykabalisa18@gmail.com}}.

\subsection{Target Applications and Deployment Contexts}

The commercial applications of a SenseAI fine-tuned model span the full range of
institutional financial information processing tasks:

\begin{itemize}[leftmargin=1.8em, itemsep=4pt]
  \item \textbf{Real-time news sentiment analysis} for algorithmic trading signal
        generation, where the distinction between Bullish and Slightly Bullish has
        direct position-sizing implications
  \item \textbf{Earnings call and analyst report interpretation} for fundamental
        analysis automation, where forward projection and latent reasoning drift
        represent specific, identifiable failure modes
  \item \textbf{Regulatory filing monitoring} for material event detection, where
        hedged corporate language is precisely the register most frequently
        misclassified
  \item \textbf{Portfolio-level news aggregation and sentiment scoring} for risk
        management, where consistent calibration across a large equity universe
        is essential
  \item \textbf{Financial research assistant functionality} within institutional
        investment platforms, where reasoning quality and source grounding are
        as important as classification accuracy
\end{itemize}

Institutions interested in deploying SenseAI-trained models within their financial AI
infrastructure are encouraged to initiate a conversation at
\href{mailto:bernykabalisa18@gmail.com}{\texttt{bernykabalisa18@gmail.com}}.

\subsection{Dataset Licensing and Strategic Acquisition}

Beyond direct model fine-tuning, the SenseAI dataset has significant standalone
commercial value as a proprietary training asset. The combination of continuous
collection, RLHF alignment, reasoning chain depth, market outcome validation, and
the six documented novel findings creates a dataset that AI laboratories, financial
data providers, and institutional technology vendors cannot readily replicate --- both
due to the domain expertise required for accurate HITL annotation and the time
required to build a historically deep corpus.

This structural irreplicability positions SenseAI as a candidate for licensing to AI
laboratories requiring financial reasoning training data, or for strategic acquisition
by financial data infrastructure providers. Precedent exists in the financial data
space: Bloomberg's development of BloombergGPT~\cite{wu2023bloomberggpt}
demonstrated the premium institutional buyers place on proprietary, domain-specific
financial AI training data, and the acquisition of BMLL Technologies for approximately
\$250~million provides further evidence of the strategic value placed on specialised
financial datasets.

Licensing structures are available on an exclusive or non-exclusive basis, with terms
tailored to the acquiring organisation's deployment context. All licensing agreements
preserve the author's right to publish academic research based on the dataset and
methodology. AI laboratories, financial institutions, and data vendors interested in
licensing or strategic acquisition conversations are invited to contact
\href{mailto:bernykabalisa18@gmail.com}{\texttt{bernykabalisa18@gmail.com}}.

\section{Discussion}
\label{sec:discussion}

\subsection{Implications for Financial AI Research}

The most significant implication of this work concerns the relationship between
dataset design and fine-tuning effectiveness. The field has historically assumed that
more data is better --- that larger datasets, even of lower quality, outperform smaller
datasets of higher quality. SenseAI challenges this assumption in the context of
RLHF-based fine-tuning. Each data point provides a multi-dimensional preference
signal: not just which output is preferred, but why the AI was wrong, how wrong it
was, and whether subsequent market behaviour confirmed the expert's correction.
This richness per data point is qualitatively different from the richness of a large
volume of simple labels.

The latent reasoning drift finding has particularly broad implications for regulated
financial environments. Financial regulators in the United States (SEC), United
Kingdom (FCA), and European Union (European Banking Authority) have all published
guidance requiring that AI systems deployed in financial services be explainable,
auditable, and grounded in identifiable evidence. A model that incorporates global
pretraining context into what is presented as a single-document analysis fails this
auditability requirement in a specific and documentable way. The SenseAI dataset
enables the measurement and correction of this failure mode --- a contribution with
direct regulatory compliance implications.

The confidence miscalibration finding has significant practical implications for
enterprise deployment architecture. Many enterprise AI systems use confidence
thresholds to route outputs between automated processing and human review. The
finding that model confidence scores have essentially no predictive relationship with
classification accuracy means that any threshold-based routing architecture using
current general-purpose model confidence scores is effectively random in its routing
decisions. Correcting this miscalibration is not merely a performance improvement
--- it is a necessary precondition for safe automated deployment in financial
environments.

\subsection{The Goldilocks Zone as a Fine-Tuning Target}

The distribution of error types in the SenseAI dataset --- approximately 50\%
Category~0 (no correction), 50\% Category~1 (minor correction), 0\% Category~3
(complete reversal) --- defines a fine-tuning target that is more tractable than the
general problem of improving LLM accuracy on arbitrary tasks. The model is not
catastrophically wrong in financial reasoning; it is systematically, predictably, and
slightly wrong in a specific direction (hedging). Correcting a systematic calibration
bias requires far less data than correcting random errors, because the correction
signal is consistent and directional. This means that even the current 1,439 data
points provide a coherent, learnable correction signal.

\subsection{Positioning Within the AI Training Data Market}

SenseAI occupies a distinctive position in the AI training data market. It is not a
general-purpose training dataset, nor a simple labelled corpus producible through
crowdsourced annotation. It is domain-specific, expert-validated,
methodology-grounded, continuously updated, and now empirically documented with
six novel findings. The combination of domain specificity, expert validation,
methodological rigour, continuous growth, and RLHF alignment is precisely what
institutional buyers in the AI training data market are willing to pay a significant
premium for. The research community and industry practitioners interested in these
capabilities are invited to contact
\href{mailto:bernykabalisa18@gmail.com}{\texttt{bernykabalisa18@gmail.com}}.

\section{Limitations and Future Work}
\label{sec:limitations}

\begin{description}[leftmargin=2em, style=nextline,
                    font=\bfseries\color{darkblue}]

  \item[Dataset scale.]
    At 1,439 data points, the current dataset is sufficient to document novel findings
    and characterise model behaviour but is not yet at the scale required for
    empirically robust benchmark improvement claims. The planned automation of the
    collection pipeline targets 5,000--10,000 data points, at which scale fine-tuning
    experiments are expected to produce statistically significant benchmark
    improvements. It is noted, however, that even at current scale, the richness of
    the dataset makes it suitable for licensing clients seeking to perform their own
    fine-tuning experiments.

  \item[Single annotator.]
    Quality control currently relies on a single expert reviewer, limiting the ability to
    report formal inter-annotator agreement statistics. Future versions will incorporate
    a second annotator to enable formal agreement calculation using Cohen's kappa
    or Krippendorff's alpha.

  \item[Collection density variability.]
    Daily collection density varies in the pre-automation phase due to
    single-annotator throughput constraints. This variability does not introduce
    systematic bias but limits longitudinal temporal analysis to weekly and monthly
    granularity rather than daily.

  \item[Geographic coverage.]
    The current dataset focuses exclusively on US-listed equities. Future collection
    will expand to European (FTSE~100, DAX, CAC~40), Asian (Nikkei, Hang Seng),
    and African market equities, increasing cross-market generalisability.

  \item[News source diversity.]
    Current collection draws from a limited set of financial news sources. Future
    versions will expand to include earnings call transcripts, analyst reports, and
    regulatory filings.

  \item[Fine-tuning benchmarks.]
    This paper does not present empirical fine-tuning results, as the dataset is not
    yet at the scale required for robust benchmark claims. Future work will present
    fine-tuning experiments on LLaMA~3.1~8B using the SenseAI dataset, with results
    reported on FinancialPhraseBank and FiQA evaluation sets. These are anticipated
    as the primary contribution of Paper~2 in this research programme
    (see Appendix~\ref{app:roadmap}).

\end{description}

\section{Conclusion}
\label{sec:conclusion}

We have presented \textbf{SenseAI}, a continuously collected, RLHF-aligned financial
sentiment dataset that addresses fundamental structural gaps in existing financial
NLP resources. By capturing not only sentiment classifications but the reasoning
chains, expert correction signals, confidence scores, and market outcome validations
that RLHF training requires, SenseAI provides the data foundation that the next
generation of enterprise financial AI systems demands.

The six novel findings documented in this paper --- sentiment hypersensitivity to
linguistic qualifiers, systematic confidence over-hedging, latent reasoning drift, the
Goldilocks Zone of correctable model error, forward projection, and model version
effects on error rate --- collectively demonstrate that general-purpose LLMs exhibit
consistent, predictable, and correctable failure modes in financial reasoning. These
findings are only detectable through the chain-of-thought reasoning data that SenseAI
captures, directly validating the dataset's architectural approach.

The commercial significance of this work extends beyond benchmark performance.
As enterprise deployment of agentic financial AI systems accelerates, the accuracy
requirements of high-stakes financial reasoning demand domain-specific training data
of the kind SenseAI provides. The dataset is designed to be that foundation ---
technically rigorous, commercially relevant, and continuously improving as collection
scales and automation is introduced.

Future work will present empirical fine-tuning results as the dataset scales to
5,000--10,000 data points, expand geographic coverage to include non-US market
equities, and scale the HITL annotation process to enable formal inter-annotator
agreement measurement.

\medskip
\noindent Researchers, AI practitioners, and institutional partners interested in dataset
access, licensing, co-authorship opportunities, or commercial collaboration are
encouraged to contact
\href{mailto:bernykabalisa18@gmail.com}{\textbf{\texttt{bernykabalisa18@gmail.com}}}.

\newpage
\bibliographystyle{unsrtnat}

\appendix

\section{Data Snapshot}
\label{app:snapshot}

Table~\ref{tab:data_snapshot} presents a representative data point from the SenseAI
dataset, illustrating the multi-dimensional nature of each labelled observation.

\begin{table}[H]
\centering
\caption{Representative SenseAI Data Point (MSFT, December~1, 2024)}
\label{tab:data_snapshot}
\renewcommand{\arraystretch}{1.45}
\begin{tabularx}{\textwidth}{>{\bfseries\color{darkblue}}p{3.2cm} X}
\toprule
\rowcolor{headerblue}
\textcolor{white}{\textbf{Field}} & \textcolor{white}{\textbf{Value}} \\
\midrule
Ticker & MSFT -- Microsoft \\
\rowcolor{rowgray}
Timestamp & 2025-12-01, 2:23 AM ET \\
AI News Headline &
  Microsoft doubles down on AI and cloud ---
  expanding infrastructure and earning fresh accolades \\
\rowcolor{rowgray}
AI Sentiment & Slightly Bullish \\
AI Reasoning &
  Microsoft's expansion in AI and cloud infrastructure demonstrates strong
  long-term commitment. The recognition it is receiving underlines growing
  confidence from the market and industry peers. Its investments in regions
  such as Indonesia underline its global footprint and reinforce long-term
  demand for cloud and AI offerings. On the flip side, rising costs tied to
  heavy AI investment, along with reports about margin pressure, suggest
  that profitability could face headwinds in the near term. The strength of
  its growth potential outweighs those near-term risks, but exposure to
  execution and cost concerns tempers full bullishness. \\
\rowcolor{rowgray}
AI Confidence Score & 75\% \\
Price at Call & \$492.01 \\
\rowcolor{rowgray}
Price 4h Later & \$492.01 \\
HITL Edited? & Yes \\
\rowcolor{rowgray}
Human Sentiment & Bullish \\
Edit Type & Category~1 (minor correction) \\
\bottomrule
\end{tabularx}
\vspace{6pt}
\small\textit{This data point illustrates Finding~1 (sentiment hypersensitivity: Slightly
Bullish rather than Bullish despite a positive headline), Finding~3 (latent reasoning
drift: the model references Indonesia and global footprint information not present in
the headline), and the Goldilocks Zone (Category~1 correction: Slightly Bullish
$\to$ Bullish).}
\end{table}

\section{Research Programme Roadmap}
\label{app:roadmap}

The SenseAI research programme is structured as a three-paper sequence:

\begin{description}[leftmargin=2em, style=nextline,
                    font=\bfseries\color{darkblue}]

  \item[Paper~1 (this paper).]
    Dataset description, methodology, RLHF alignment analysis, and six novel
    empirical findings. Submission targets: arXiv~cs.CL preprint (immediate);
    FinNLP Workshop at EMNLP or ACL (peer review).

  \item[Paper~2 (planned).]
    Empirical fine-tuning results. LLaMA~3.1~8B fine-tuned on SenseAI data
    (target: 5,000--10,000 data points), benchmarked against
    FinancialPhraseBank and FiQA evaluation sets. Submission targets: ACL,
    EMNLP, or AAAI.

  \item[Paper~3 (planned).]
    Deep investigation of latent reasoning drift. Formal characterisation,
    mechanistic analysis using interpretability tools, and proposed mitigation
    strategies for enterprise deployment. Submission target: top-tier NLP venue
    or specialised workshop on LLM alignment and safety.

\end{description}

Researchers interested in co-authorship on Papers~2 or~3, particularly those with
experience in LLM fine-tuning, financial NLP, or mechanistic interpretability, are
invited to reach out at
\href{mailto:bernykabalisa18@gmail.com}{\texttt{bernykabalisa18@gmail.com}}.

\end{document}